\documentclass[runningheads]{llncs}
\usepackage[T1]{fontenc}
\usepackage{graphicx}
\usepackage{amsfonts}
\usepackage{footnote}
\makesavenoteenv{tabular}
\makesavenoteenv{table}

\usepackage{array}
\newcolumntype{P}[1]{>{\centering\arraybackslash}p{#1}}
\begin{document}
\title{Deep Neural Networks for Encrypted Inference with TFHE}
%
%
\author{Andrei Stoian \and Jordan Frery \and  Roman Bredehoft \and Luis Montero \and 
Celia Kherfallah \and 
Benoit Chevallier-Mames}
\authorrunning{A. Stoian et al.}

%
\institute{Zama \footnote{
\email{hello@zama.ai} \url{http://zama.ai}} }
\maketitle              


\begin{abstract}
    
Fully homomorphic encryption (FHE) is an encryption method that allows to perform computation on encrypted data, without decryption. FHE preserves the privacy of the users of online services that handle sensitive data, such as health data, biometrics, credit scores and other personal information. A common way to provide a valuable service on such data is through machine learning and, at this time, Neural Networks are the dominant machine learning model for unstructured data. 

In this work we show how to construct Deep Neural Networks (DNN) that are compatible with the constraints of TFHE, an FHE scheme that allows arbitrary depth computation circuits. We discuss the constraints and show the architecture of DNNs for two computer vision tasks. We benchmark the architectures using the Concrete stack\footnote{https://github.com/zama-ai/concrete-ml}, an open-source implementation of TFHE.

\end{abstract}



%
%
\section{Introduction}

Neural Networks (NNs) are machine learning (ML) models that have driven the recent expansion of the field of Artificial Intelligence (AI). Their performance on unstructured data such as images, sound and text is unmatched by other ML techniques. Moreover, deep NNs obviate the need for complex feature engineering and process raw data directly, making them easier to deploy in production. Applications of NNs include image classification, face recognition, voice assistants, and search engines, tools which today are a staple of the user experience online. Deployment of such models in SaaS applications raises a security risk: they are a target of malevolent entities that seek to steal the sensitive user data these models process. 

Privacy-preserving technologies, such as multi-party computing (MPC) and fully homomorphic encryption (FHE), provide a solution to the risk of data leaks, eliminating it by design. Notably, FHE encrypts user data and allows a third party to process the data in its encrypted form, without needing to decrypt it. Only the data owner can decrypt the result of the computation. Thus, an attacker can only steal encrypted data they can not decrypt. 

In this work we show how to build neural networks that are FHE compatible, while minimizing the cryptography knowledge needed by the machine learning practitioner. We based our work on the Concrete Library \cite{WAHC:CJLOT20} which uses TFHE \cite{Chillotti2019TFHEFF}, works over integers, provides a fast \emph{programmable} bootstrapping mechanism, and performs exact computation. 

\section{Related work}

Several alternative approaches exist for neural network inference over encrypted data. All use NNs with integer weights and activations and many of them rely on "leveled" fully homomorphic encryption schemes that do not use bootstrapping, such as CKKS \cite{CKKS} and YASHE \cite{YASHE}. 

CryptoNets \cite{pmlr-v48-gilad-bachrach16} uses YASHE which supports the computation of polynomials of encrypted values. CryptoNets are NNs quantized to integers (of 5-10 bits) with activation functions expressed as low-degree polynomials. CryptoNets achieve 99\% accuracy on MNIST using a three layer network with an inference time of 570 seconds/image. 

FHE-DiNN \cite{DINN}, a TFHE based approach, quantizes inputs, intermediate values and weights to binary values. In this case, the training is done with \texttt{hardSigmoid} activation which is swapped for the \texttt{sign} function in inference. However, binary NNs are hard to train and do not perform well in many ML tasks such as object detection and speech processing.

Another TFHE approach, SHE \cite{SHE}, uses bit series representation of encrypted values and boolean gates. They run NNs that fit within a maximum multiplicative depth budget and, by avoiding expensive multi-bit PBSs, they achieve inference of a ShuffleNet on ImageNet with a latency of 18 000 seconds/image. They rely on logarithmic quantization of weights which allows to reduce multiplicative depth for the convolution layers by using bit-shifts. Sums, \texttt{relu} and \texttt{maxpool} are computed using boolean gates. 

Leveled approaches such as SHE and CryptoNets are  limited by the maximum multiplicative depth budget, which, in turn, limits the supported network types and their depth. Moreover, some schemes such as CKKS  are approximate by design, as the noise corrupts some of the message bits.

In this work we propose an approach to train arbitrary NNs which can have any depth, number of neurons and activation functions. Furthermore, our approach performs exact computation in FHE: the noise of the encryption scheme does not corrupt the values that are processed. Thus results in FHE are the same as in the clear - there is no degradation of accuracy when moving to encrypted inference - which is a major advantage when putting models in production.

\section{Neural Network Training for Encrypted Inference}

Training NNs is usually done in floating point, but most FHE schemes, including TFHE, only support integers. Consequently, quantization must be used, and two main approaches exist: 
\begin{enumerate}
    \item Post-training quantization is commonly used \cite{pmlr-v48-gilad-bachrach16,SHE}, but, in this mode, NNs lose accuracy when the quantization bit-width is lower than 7-8 bits. With per-channel quantization, or logarithmic quantization, which are more complex to implement, as few as 4 bits were used for weights and activations without loss of accuracy \cite{PERCHANNEL}.  
    \item Quantization-aware Training (QAT), used in this work and in \cite{DINN}, is an approach that adds quantizers to network activations and weights during training. QAT enables extreme quantization with less than 4 bit weights and activations.
\end{enumerate}

To support arbitrarily deep NNs and any activation function, we make use of the programmable boostrapping mechanism \cite{PBS} (PBS) of TFHE. PBS reduces the noise in  accumulators of ciphertext leveled operations (addition, multiplication with clear constants) but also allows to apply a lookup-table (TLU) on its input ciphertext.

The TFHE PBS mechanism has a rather high computational cost, and this cost depends on the number of bits of the encrypted value to be boostrapped. It is convenient to keep the accumulator size low, in order to speed up the PBS computation. However, reducing accumulator bit-width has a negative impact on network prediction performance, so a compromise needs to be found.

We describe here a QAT strategy that can process all the intermediate encrypted values as integers. In this way, training an FHE compatible network becomes purely a machine learning problem and no cryptography knowledge is needed by the practitioner. To build a TFHE compatible NN, the constraints on the network architecture are the following:

\begin{itemize}
    \item All layers that sum or multiply two encrypted values, such as convolution \texttt{conv} and fully-connected \texttt{fc}, must have quantized inputs. This is easily achieved using QAT frameworks.
    \item The bit-width of the accumulators of layers such as \texttt{conv}, \texttt{fc} must be bounded. To achieve this, we use pruning.
\end{itemize}

\begin{figure}
    \centering
    \includegraphics[width=0.5\textwidth]{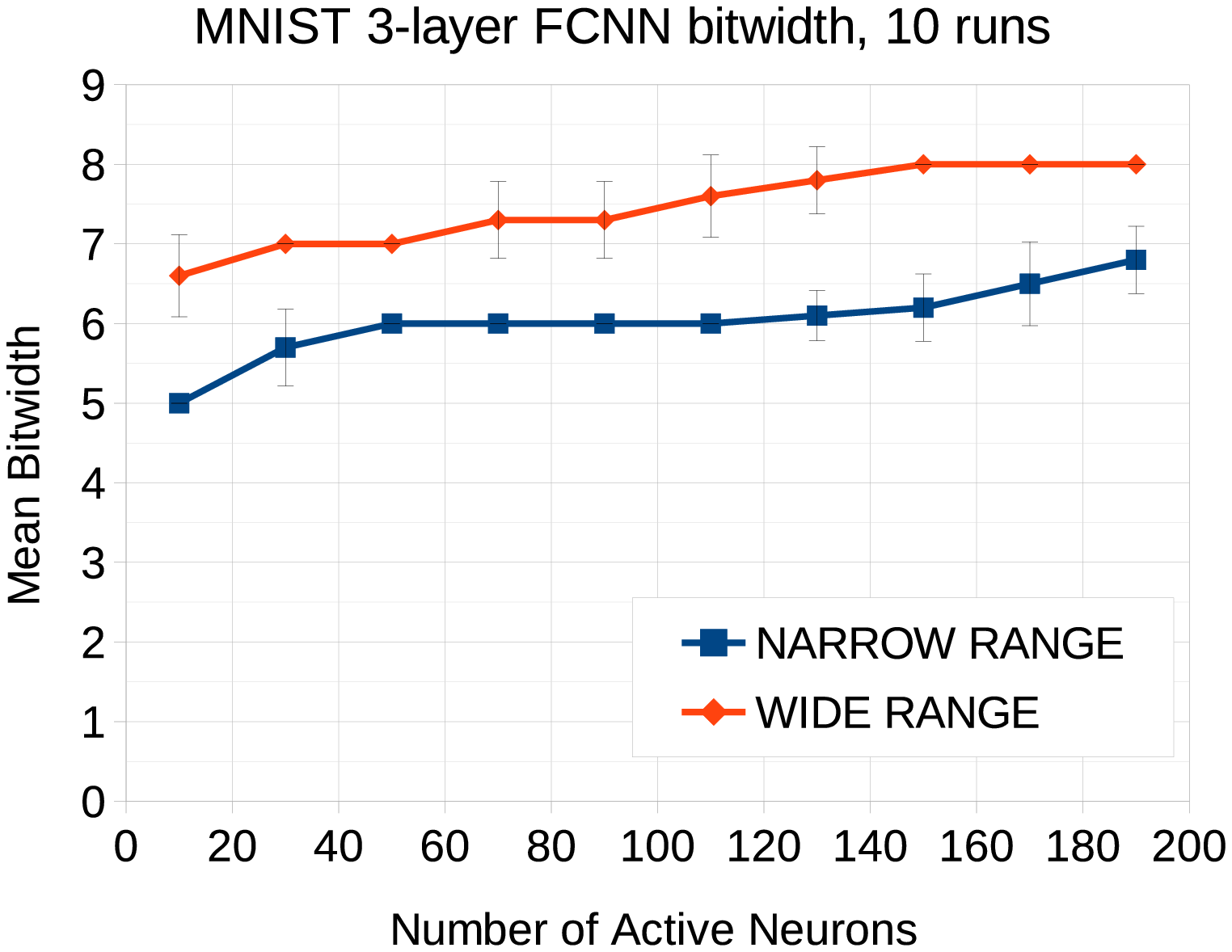}
    \includegraphics[width=0.45\textwidth]{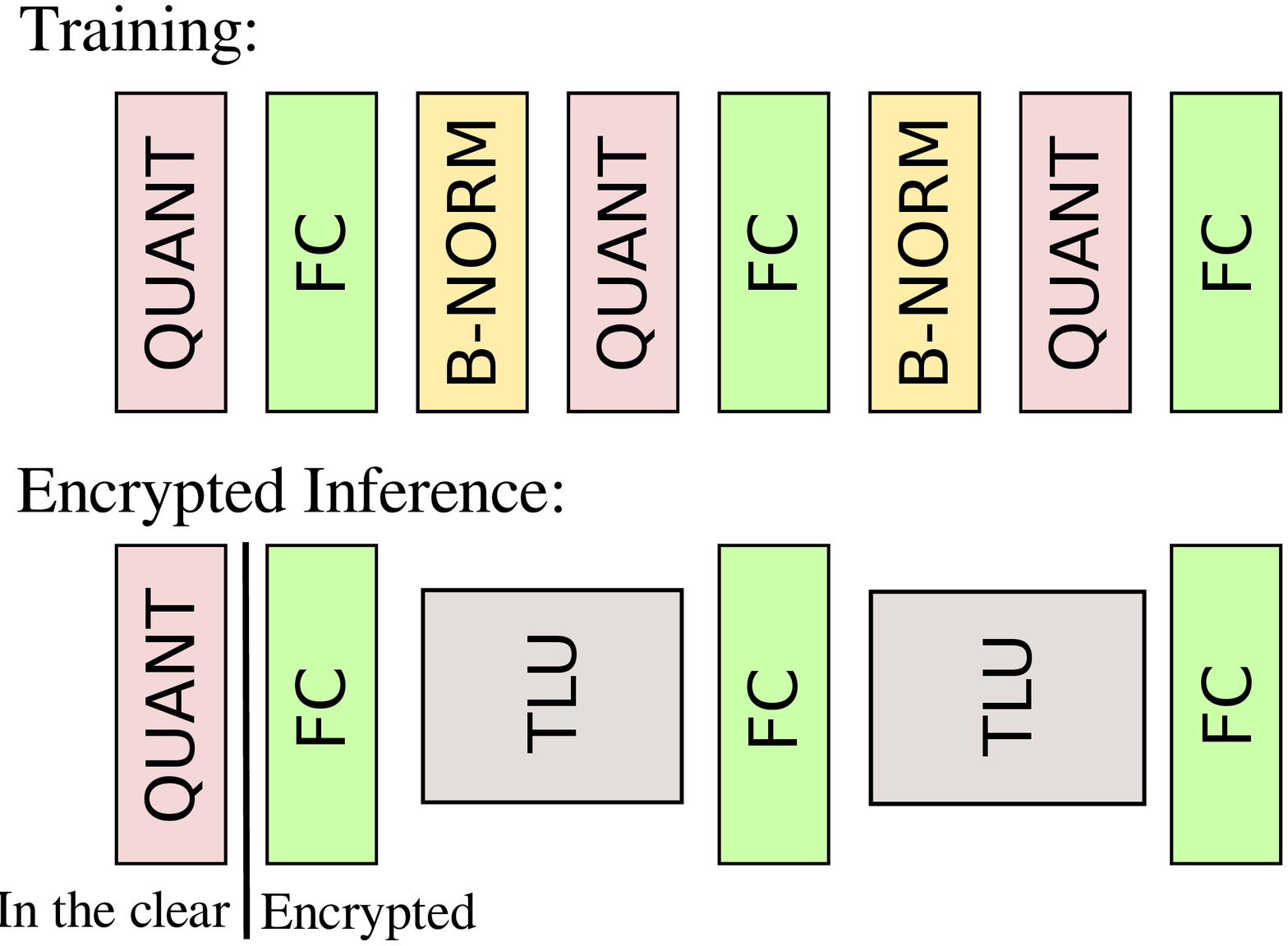}
    \caption{Left: accumulator size while varying the number of active neurons during pruning for a 3-layer fully-connected network with 2 bit weights and activations. Two quantization modes are shown: Narrow range uses values $[-2^{b-1}+1,2^{b-1}-1]$, while Wide range uses $[-2^{b-1}, 2^{b-1}-1]$. Right: the structure of a 2 layer convolutional network in training and during inference. Univariate layers are fused to table-lookups, implemented with PBS.}
    \label{fig:pruning_arch}
\end{figure}

To control the accumulator bit-width while keeping the training dynamics stable, we use $L^1$-norm unstructured pruning. Figure \ref{fig:pruning_arch} shows the impact of pruning on the accumulator size for two quantization modes: narrow and wide range.

While the inputs of \texttt{conv} and \texttt{fc} layers need to be quantized, it is possible to use floating point layers for all univariate operations such as batch normalization, quantization, and activations. 

In our FHE compatible NNs the outputs of a \texttt{conv} or \texttt{fc} are processed by a sequence of univariate operations that ends with quantization. This sequence of functions takes integers and has integer outputs, but the intermediary computations in these operations can use float parameters. Thus, batch normalization, activation functions, neuron biases and any other univariate transformation of \texttt{conv} or \texttt{fc} outputs does not need quantization. Figure \ref{fig:pruning_arch} shows the architecture of the network during training and inference. 

\section{Neural Network Inference using TFHE}

Inference of our FHE NNs is based on quantized implementations of NN operators that add or multiply together encrypted values. Convolutional, fully connected and average pooling layers use the quantized formulation from \cite{8578384}. Since uniform quantization is used, we can define a quantized value $r$ as $r = S(q- Z)$ where $S$ is the quantization scale, $Z$ is the quantization zero-point and $q$ is the integer representation of the value. Next, the fully connected layer, with inputs $x$, weights $w$ and outputs $o$, with per-tensor quantization parameters $(S_x,Z_x)$, $(S_w,Z_w)$  can be written as:

\begin{equation}
    S_o(q_o^{k} - Z_o) = \sum_{i=0}^N S_x(q_x^{i} - Z_x)S_w(q_w^{(i,k)} - Z_w) + b^k
\label{eq:fcq}
\end{equation}

\noindent where $k$ is the index of a neuron in the layer and $N$ is the number of connections of the neuron and $b^k$ is the bias of the $k$-th neuron. A convolutional layer can be expressed by extending the sum to the height, width and channel dimensions. Equation \ref{eq:fcq} can be re-written to separate integer and floating point computations (note that zero-points $Z_x,Z_o,Z_w$ are integers). 

\begin{equation}
    q_o^{k} = b^k + Z_o + \frac{S_xS_w}{S_o}\sum_{i=0}^N(q_x^i - Z_x)(q_w^{(i,k)} - Z_w)
\label{eq:fcq}
\end{equation}

Therefore we can separate the equation in a floating point univariate function $f$ and a sum over products of encrypted inputs and clear weights:

\begin{equation}
    q_o^{k} = f(\Sigma ) ~~\textrm{where}~~ f(q) = b^k + Z_o + \frac{S_xS_w}{S_o}q ~~\textrm{and}~~ \Sigma = \sum_{i=0}^N(q_x^i - Z_x)(q_w^{(i,k)} - Z_w)
\label{eq:fcq_sep}
\end{equation}

The univariate function $f$ in eq. \ref{eq:fcq_sep} takes integer inputs. We compose this function with the batch-normalization, and, finally, with the quantization function $Q(x) = floor\left(\frac{x}{S_x}\right) + Z_x$. Thus $f$ becomes a function defined on $\mathbb{Z}$, with values in $\mathbb{Z}$ and can be implemented as a lookup table with a PBS in FHE, without any loss of precision.

The complete NN computation can now be expressed over integers using the following operations: multiplication of an encrypted value and a clear constant, sums of encrypted integer values, table lookup of encrypted integer values. In our implementation of TFHE, Concrete, we encode integers in two different ways: integers up to 8 bits are encoded into a single ciphertext, and integers between 9-16 bits are encoded with a CRT representation into several ciphertexts as described in  \cite{ZAMA_CRT_OPT}. This contrasts to previous works, such as \cite{SHE}, that encode each bit of an integer as an individual ciphertext and use boolean gates to build arithmetic circuits.

An automated optimization process \cite{ZAMA_CRT_OPT} determines the cryptographic parameters of the circuit, based on several factors: (1) the \emph{circuit bit-width}, defined as the minimum bit-width necessary to encode the largest integer value obtained anywhere in the NN's integer-based evaluation, (2) the maximum 2-norm of the integer weight tensors of the layers, and (3), the desired probability of error of the PBS.  The optimization process determines the cryptosystem parameters (LWE dimension, polynomial size, GLWE dimension, etc.) to ensure a fast execution, the target probability of failure and the security level (using the lattice-estimator \cite{cryptoeprint:2015/046}). We set the PBS error probability sufficiently low to ensure full correctness of the results, i.e. the results in the clear are always the same as those in FHE, up to a user-defined error-rate, e.g. $10^{-6}$, for one full NN inference.

\section{Experimental Results}

The networks were implemented in PyTorch with Brevitas \cite{brevitas} and converted to FHE with Concrete-ML \cite{concrete-ml}. We ran experiments on two datasets with several neural network architectures, in two quantization modes (see Figure. \ref{fig:pruning_arch}, left). The test machine had an Intel i7-11800H CPU with 8 cores and we used 16 threads for the experiments. 

\begin{table}
\caption{Experimental results obtained with Brevitas and Concrete-ML}\label{tab1}
\begin{tabular}{|P{2.75cm}|P{1.25cm}|P{1.5cm}|P{1.5cm}|P{1.5cm}|P{1.5cm}|P{2cm}|P{2cm}|}
\hline
Network & 
Quant. bits & 
Active Neurons & 
Narrow range &
Data-set & 
Circuit bit-width & 
Accuracy & 
Inference time (s)\\
\hline 
3-layer FCNN\footnote{Three FC layers with 192, 192 and 10 neurons} & 2/2 & 150 & Yes & MNIST &  6 & 92.2\% & 31 \\
3-layer FCNN & 2/2 & 90 & No & MNIST &  7 & 96.5\% & 77 \\
3-layer FCNN & 2/2 & 190 & No & MNIST &  8 & 97.1\% & 300 \\
LeNet & 2/2 & 190 & No & MNIST &  \centering  8 & 97.6\% & 2780 \\
6-layer CNN\footnote{Four conv layers with 8,8,16 and 16 filters followed by a FC layer with 120 neurons and a final FC layer for classification} & 2/2 & 190 & No & \centering  MNIST & 8 & \centering 98.7\% & 5072 \\
VGG-9\footnote{Six conv layers with: 64, 64, 128, 128, 256, 256 filters, followed by two 512 neuron FC layers and a final FC layer for classification} & 2/2\footnote{Inputs are quantized in 8 bits, but all other activations use 2 bits} & all & Yes &  CIFAR10 &  \centering  13 & 87.5\% & 18000\footnote{Estimated time for a 8 core machine, using 16 threads} \\

\hline
\end{tabular}
\end{table}

\vspace{-0.5cm}
\section{Conclusion}

Our approach to encrypted inference for Neural Networks shows several advantages over other methods. First, we believe our method is easier to use than other works, since the problem of making an FHE compatible network becomes strictly an ML problem and no cryptography knowledge is needed. Second, the computations in FHE are correct with respect to the computations in the clear and, using TFHE, noise does not corrupt the encrypted values. Thus, once a network is trained incorporating the quantization constraints, the accuracy that is measured on clear data will be the same as that on encrypted data. Finally, our approach, using PBS, shows competitive accuracies in FHE and allows to convert arbitrary depth networks using any activation function to FHE. Networks up to 9 layers were shown, but deeper NNs can easily be implemented.  

Preliminary code for the MNIST classifier is available\footnote{\texttt{https://github.com/zama-ai/concrete-ml/tree/release/0.5.x/use\_case\_examples}} and code for the CIFAR10 classifier will be released soon.

Many possible strategies can be employed to improve upon this work, in order to support larger models, such as ResNet, on larger data-sets like ImageNet. For example, a better pruning strategy could decrease the PBS count, per-channel quantization can improve accuracy, and faster step functions in FHE could improve the overall speed. 

\bibliographystyle{splncs04}
\bibliography{biblio}

\end{document}